\documentclass{article}
\usepackage{spconf,amsmath,graphicx}
\usepackage{multirow}
\usepackage[table,xcdraw]{xcolor}
\usepackage{epsfig}
\usepackage{graphicx}
\usepackage{subcaption}

\title{Deep feature fusion for self-supervised monocular depth prediction}
\name{Vinay Kaushik, Brejesh Lall}
\address{Dept. of Electrical Engineering, IIT Delhi}
\begin{document}
%
\maketitle
\begin{abstract}
Recent advances in end-to-end unsupervised learning has significantly improved the performance of monocular depth prediction and alleviated the requirement of ground truth depth. Although a plethora of work has been done in enforcing various structural constraints by incorporating multiple losses utilising smoothness, left-right consistency, regularization and matching surface normals, a few of them take into consideration multi-scale structures present in real world images. Most works utilize a VGG16 or ResNet50 model pre-trained on ImageNet weights for predicting depth. We propose a deep feature fusion method utilizing features at multiple scales for learning self-supervised depth from scratch. Our fusion network selects features from both upper and lower levels at every level in the encoder network, thereby creating multiple feature pyramid sub-networks that are fed to the decoder after applying the CoordConv solution\cite{liu2018intriguing}. We also propose a refinement module learning higher scale residual depth from a combination of higher level deep features and lower level residual depth using a pixel shuffling framework that super-resolves lower level residual depth. We select the KITTI dataset\cite{geiger2013vision} for evaluation and show that our proposed architecture can produce better or comparable results in depth prediction.
\end{abstract}
\begin{figure*}
\begin{center}
\includegraphics[width=\textwidth]{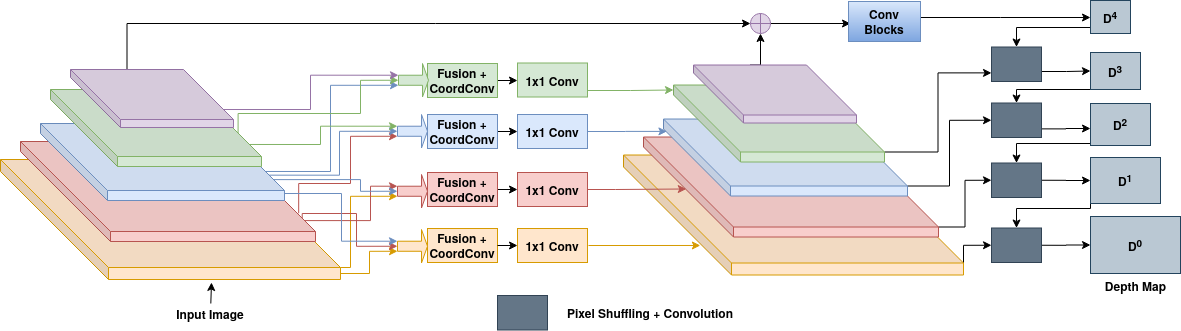}
\end{center}
\caption{Architecture for deep feature fusion encoder-decoder with depth refinement for predicting self-supervised depth.} \label{fig1}
\end{figure*}

\section{Introduction}
\label{sec:intro}
Depth estimation is a fundamental problem of computer vision encompassing wide range of applications in areas such as robotics, augmented reality, action recognition, human pose estimation and scene interaction. Traditional depth predicting algorithms rely of certain assumptions like multi-view camera data, structure from motion or stereopsis for efficiently estimating scene depth. To alleviate this problem, recent deep learning approaches pose depth prediction as a supervised learning problem. These methods depend on large collections of ground truth depth information for training a deep learning model to predict per pixel depth. While these methods work well, there are many scenarios where there is scarce possibility of getting ground truth depth data. 

Self-supervised depth prediction solves the depth prediction problem by tackling the problem of image view synthesis. Garg et al.\cite{garg2016unsupervised} proposed a deep network that enforced a photometric constraint on stereo image data for predicting depth. DVSO \cite{yang2018deep} uses sparse depth computed from visual odometry pipeline as a supervisory signal for improving depth prediction. Although a lot of work has been done on enforcing multiple constraints on learning depth, like surface normal constraint, left-right consistency constraint, etc. by constructing a loss for the same, not much has been done to enforce changes to the encoder-decoder architecture itself, that has to actually learn those constraints. The most crucial of them is capturing multi-scale structural information present in the scene itself. 

In this work, we pose depth prediction as an unsupervised problem, where our model learns to predict multi-scale depth by learning pixel level correspondence between rectified pairs of stereo images with known baseline. Our work optimises the depth prediction pipeline by utilising a novel deep learning framework bringing critical architecture improvements over existing methods enabling us to achieve better accuracy and qualitative improvements in our results.  
In summary, we make the following contributions: (i) A self-supervised learning architecture utilising multiple feature pyramid sub-networks for feature fusion at every feature scale, (ii) Our model takes advantage of the CoordConv solution\cite{liu2018intriguing} by concatenating coordinate channels centered at principal point in every FPN sub-network along with every skip connection, (iii) A residual refinement module that leverages lower resolution residual depth utilizing a residual sub-network, incorporating pixel shuffling for super-resolving predicted depth, (iv) The proposed method achieves state-of-the-art performance on KITTI driving dataset\cite{geiger2013vision} and further improves the visual quality of the estimated depth maps. 

\section{Related Work}
\label{sec:related}
Predicting depth from images has always been a crucial task in computer vision. Majority of traditional approaches use stereo or multi-view images to predict depth. Advent of deep learning led to collecting depth ground truth by expensive scanners, posing depth prediction as a supervised learning problem. We target on a specific domain of monocular depth prediction, posing depth prediction as a self-supervised learning problem, where given only single image as input, we aim to predict it's depth without considering any scene prior. 

\subsection{Supervised Depth Prediction}
Laina et al.\cite{laina2016deeper} introduced ResNet based FCN architecture for predicting supervised depth. Hu et al.\cite{hu2019revisiting} constructed a multi-scale feature fusion module to produce better depth at object edges. Chen et al.\cite{chen2019structure} formulated a structure-aware residual pyramid network for learning supervised depth.
\subsection{Self-supervised Monocular Depth Prediction}
Garg et al.\cite{garg2016unsupervised} presented the problem of predicting depth as a learning problem with depth prediction as an intermediate step of image synthesis. The photometric error is computed to train the deep network. Godard \cite{godard2017unsupervised} proposed that the depth produced by both left and right images must be consistent with each other and constructed a left-right consistency loss enforcing the same. He also utilized an effective weighted combination of SSIM and L1 loss along with a smoothness constraint to further optimize the predicted depth. Pillai \cite{pillai2019superdepth} introduced depth super resolution by using a sub-pixel convolution layer thereby improving depth at higher resolutions. UnDispNet\cite{kaushik2019undispnet} provided a cascaded residual framework utilising super-resolution for refining depth by utilising multiple autoencoders.

Almalioglu et al.\cite{almalioglu2019ganvo} utilised a generative adversarial network by assuming the depth encoder as a generator network and feeding the synthesized image to the discriminator network. Feng et al.\cite{feng2019sganvo} constructed a stack of GAN layers, with higher layers estimating spatial features and lower layers learning depth and camera pose. Zhou\cite{zhou2017unsupervised} presented an algorithm to learn depth from a monocular video. His framework comprised of separate depth and pose networks, with the loss constructed by warping temporal views by combining camera pose with the predicted depth of the target image. GeoNet\cite{yin2018geonet} resolved texture ambiguities by incorporating an adaptive geometric consistency loss. Godard\cite{godard2019digging} proposed a shared encoder framework for predicting depth and pose from monocular videos. He also introduced a minimum photometric error loss that learnt optimal depth at every pixel from a set of temporal frames. UnDeepVO\cite{li2018undeepvo} combined both temporal and spatial constraints of photometric consistency and depth consistency to create a single framework for egomotion and depth estimation. GLNet\cite{chen2019self} designed a loss to only learn depth from static regions of the image. GLNet also proposed an online refinement strategy that further improved depth prediction and thereby provided an efficient solution for better generalization of the learnt depth on novel datasets. 



\begin{figure}
\centering
        \begin{subfigure}[b]{0.20\textwidth}
                \centering
                \includegraphics[width=\linewidth]{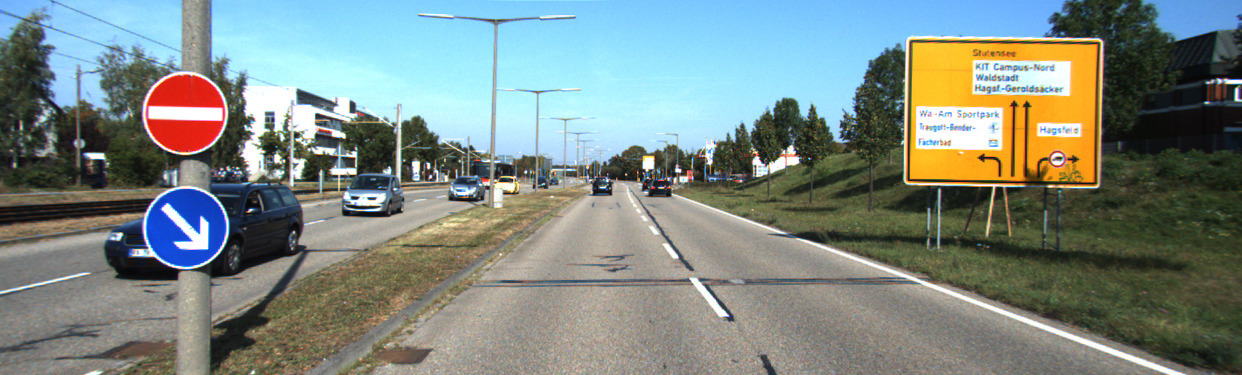}
        \end{subfigure}
        \begin{subfigure}[b]{0.20\textwidth}
                \centering
                \includegraphics[width=\linewidth]{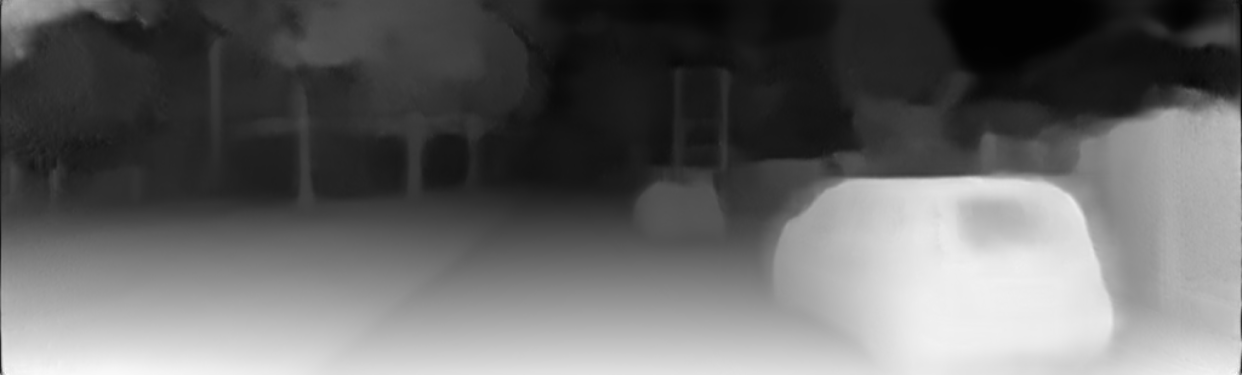}
        \end{subfigure}
        \begin{subfigure}[b]{0.20\textwidth}
                \centering
                \includegraphics[width=\linewidth]{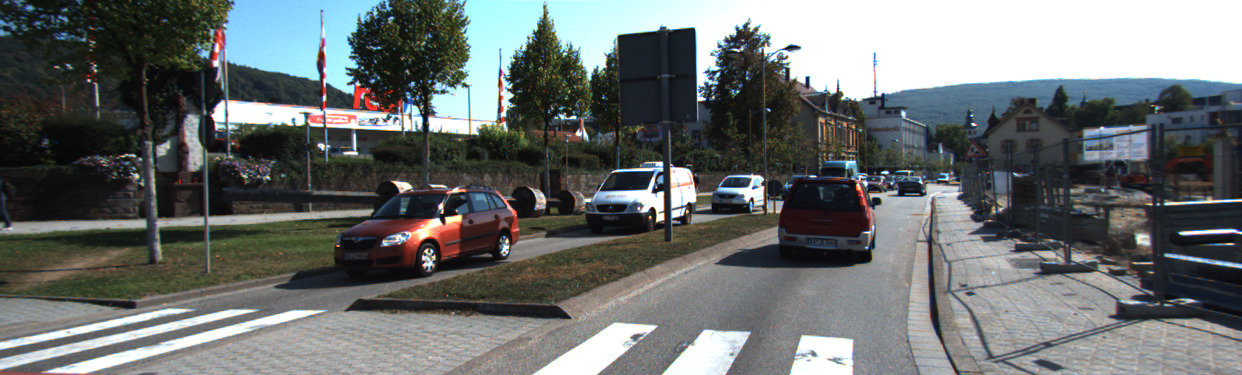}
        \end{subfigure}
        \begin{subfigure}[b]{0.20\textwidth}
                \centering
                \includegraphics[width=\linewidth]{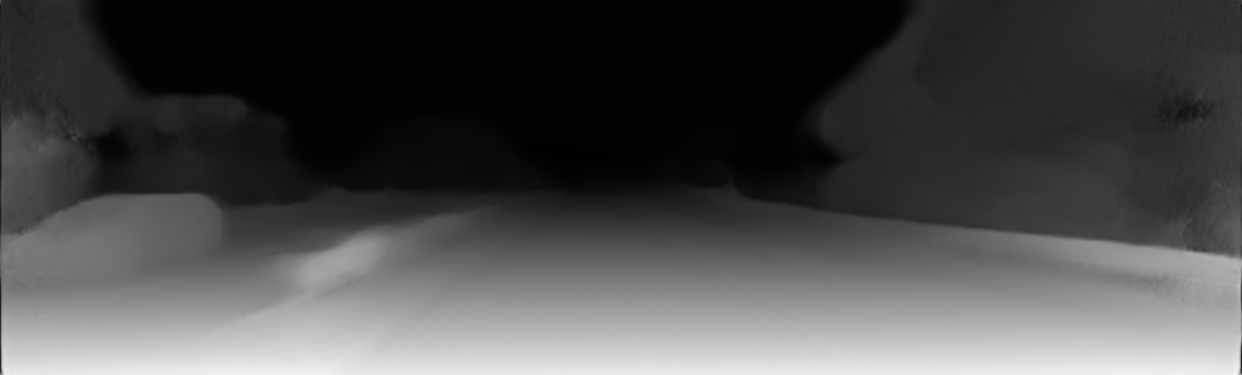}
        \end{subfigure}
        \begin{subfigure}[b]{0.20\textwidth}
                \centering
                \includegraphics[width=\linewidth]{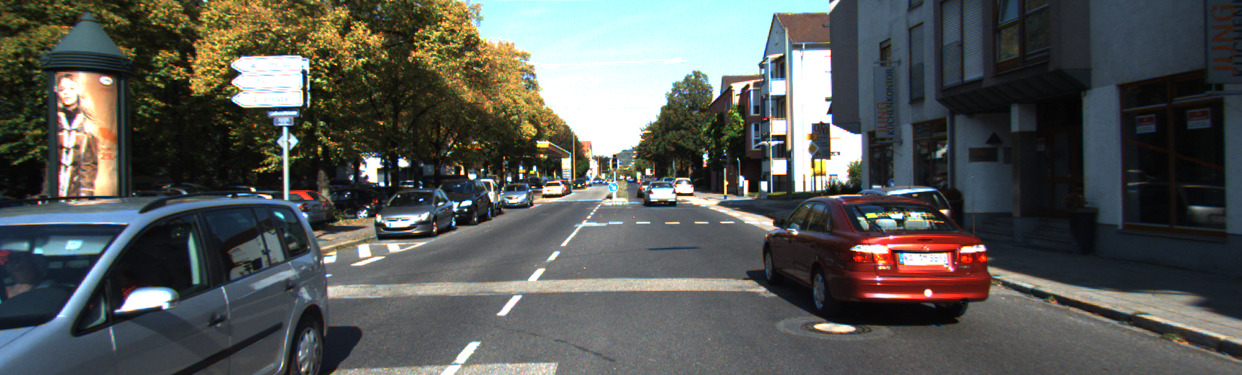}
        \end{subfigure}
        \begin{subfigure}[b]{0.20\textwidth}
                \centering
                \includegraphics[width=\linewidth]{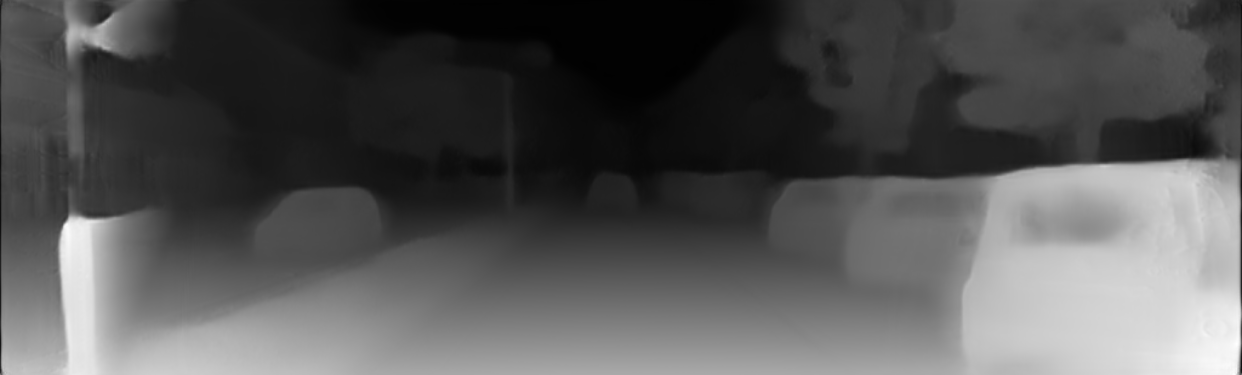}
        \end{subfigure}
        \begin{subfigure}[b]{0.20\textwidth}
                \centering
                \includegraphics[width=\linewidth]{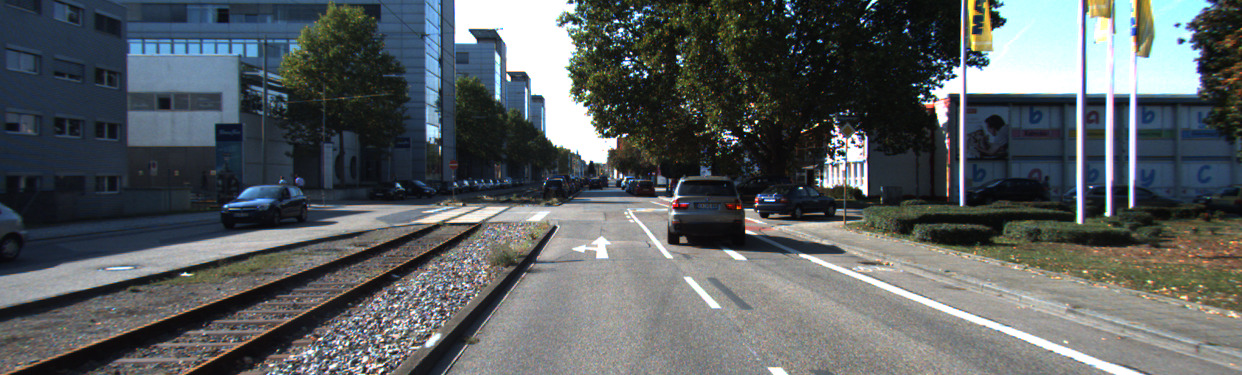}
        \end{subfigure}
        \begin{subfigure}[b]{0.20\textwidth}
                \centering
                \includegraphics[width=\linewidth]{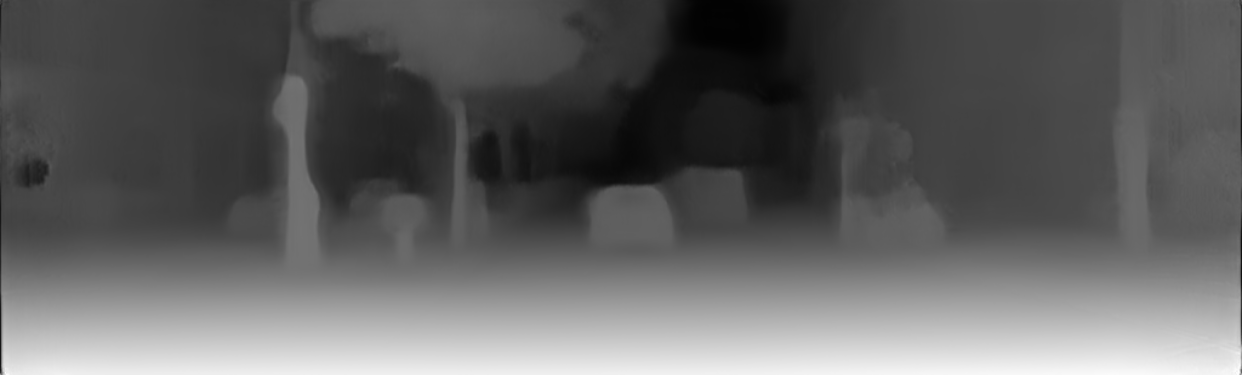}
        \end{subfigure}
\caption{Qualitative results KITTI diving dataset 2015\cite{geiger2013vision}}
\label{fig:fig2}
\end{figure}

\begin{table*}
\begin{tabular}{|c|p{1.0cm}|p{0.8cm}|p{0.8cm}|p{1cm}|p{1cm}|c|c|c|p{1cm}|c}
\cline{1-10}
Method           & Dataset & \cellcolor[HTML]{CBCEFB}Abs Rel & \cellcolor[HTML]{CBCEFB}Sq Rel & \cellcolor[HTML]{CBCEFB}RMSE & \cellcolor[HTML]{CBCEFB}RMSE log & \cellcolor[HTML]{CBCEFB}\textit{D1-all} & \cellcolor[HTML]{9B9B9B}$\delta$\textless$1.25$ & \cellcolor[HTML]{9B9B9B}$\delta$\textless$1.25^2$ & \cellcolor[HTML]{9B9B9B}$\delta$\textless$1.25^3$ &                                          \\ \cline{1-10}
Deep3D\cite{xie2016deep3d}  & K       & 0.412                           & 16.37                          & 13.693                       & 0.512                            & 66.85                                   & 0.69                                      & 0.833                                       & 0.891                                       &                                          \\ \cline{1-10}
Deep3Ds\cite{xie2016deep3d} & K       & 0.151                           & 1.312                          & 6.344                        & 0.239                            & 59.64                                   & 0.781                                     & 0.931                                       & 0.976                                       & \multirow{-3}{*}{}                       \\ \cline{1-10}
Godard\cite{godard2017unsupervised} pp          & K       & 0.124                           & 1.388                          & 6.125                        & 0.217                            & 30.272                                  & 0.841                                     & 0.936                                       & 0.975                                       & \cellcolor[HTML]{CBCEFB}\begin{scriptsize}Lower is better\end{scriptsize}  \\ \cline{1-10}
UnDispNet\cite{kaushik2019undispnet} pp            & K       & 0.110                           & 1.059                          & 5.571                        & 0.195                            & 28.656                                  & 0.858                                     & 0.947                                       & 0.980                                       & \cellcolor[HTML]{9B9B9B}\begin{scriptsize}Higher is better\end{scriptsize} \\ \cline{1-10}
Ours no pp            & K       & 0.107                          & 1.056                          & 5.370                        & 0.188                            & 27.379                                  & 0.866                                     & 0.955                                      & 0.983                                       &                                          \\ \cline{1-10}
Ours no fusion            & K       & 0.108                          & 1.098                         & 5.425                        & 0.187                           & 27.218                             & 0.869                                  & 0.953                               & 0.981                                 &                                          \\ \cline{1-10}
Ours no CoordConv            & K       & 0.106                          & 1.082                         & 5.371                        & 0.185                           & 26.727                                 & 0.870                                  & 0.955                               & 0.983                                  &                                          \\ \cline{1-10}
\textbf{Ours}            & \textbf{K}       & \textbf{0.104}                          & \textbf{1.022}                          & \textbf{5.290}                        & \textbf{0.185}                           & \textbf{26.163}                                 & \textbf{0.871}                                  & \textbf{0.956}                                 & \textbf{0.984}                                    &                                          \\ \cline{1-10}
\end{tabular}
\caption{Results on the KITTI 2015 Stereo 200 training set disparity images\cite{geiger2013vision}.For training, K is the KITTI dataset\cite{geiger2013vision}. pp is the disparity post processing specified by \cite{godard2017unsupervised}.}
\label{table:tab1}
\end{table*}

\section{METHODOLOGY}
\label{sec:methedology}
This section describes the working of our self-supervised depth prediction framework in detail. We introduce a deep feature fusion method, combining features at multiple scales to compute stereo depth without requiring any supervision or ground truth depth. We enforce the CoordConv solution\cite{liu2018intriguing} at critical places in our network to optimise scale awareness in our network. We also describe the residual refinement module for learning higher scale residual depth. 

\subsection{Depth Prediction as View Reconstruction}
Our network considers predicting fine grained depth as learning a dense correspondence field that shall be applied to the source image for reconstructing the target image of the scene. In a stereo setup, source and target images are the left and right images respectively. Our networks takes left image $I_l$ as input and predicts depth $D_l$. Given the left depth $D_l$ and right image $I_r$, we can generate $\hat{I_l}$ as the reconstructed left image. Same can be done with the right depth as input. We can simply compute $\hat{d}=bf/d$  as the metric depth, given baseline $b$ and focal length $f$.  
We compute depth of both left and right images, at 4 scales for training our network.

\subsection{Feature Fusion Network}
The core of our architecture is the Feature Fusion Network. Our pyramidal encoder utilises a combination of high level and lower level features, learning global image structure as well as the intricate shape details present in the scene. Inspired from \cite{chen2019structure}, our fusion network takes input as features at multiple scales, fuses them based on their neighbourhood and generates a richer set of features as shown in Fig\ref{fig1}. Unlike \cite{chen2019structure}, we don't reshape and feed all the features at all scales, and make sure that the feature at a given scale is given more importance than the feature at a lower or higher scale by reserving more channels for the feature at given scale. The encoder extracts a set of L features $\{F^p_E\}^L_{p=1}$, where $p$ is the feature level. The feature shape decreases by 2 at every level as the level increases due to convolution with stride 2. 
Our feature fusion network computes $FF^p_D$ taking input as a combination of upsampled features.
\begin{equation}
FF^p_D = \{F^{i,p}_{UP}\}_{i=p-1}^{p+1},
\end{equation}
where $F^{i,p}_{UP}$ is the encoded feature at level $i$ reshaped to feature at level $p$. A convolution operation with stride 1 is applied over the set of fused features to generate a set of richer features at every scale. These features form a new set of encoded features that are then forwarded to the CoordConv block before being send to decoder module.

\subsection{The CoordConv Solution}
The CoordConv block\cite{liu2018intriguing} creates three additional channels for every set of feature it receives. These coordinate channels are simply concatenated to the incoming representations. These channels contain hard-coded coordinates which is one channel for the $i^{th}$ coordinate, one for the $j^{th}$ coordinate and one for the polar(radius) coordinate defined as $r=\sqrt{(i - h/2)^2+ (j - w/2)^2}$ .
These channels are created for every feature set before being fed to the feature fusion network. Also, as a standard, these blocks are applied to the skip connections in the fusion encoder-decoder module. 
\subsection{Residual Refinement Module}
Our residual refinement module takes input as fused features $FF^p_E$ generated by the convolutional decoder, which is then fed to a set of convolutional features to compute residual depth. This residual depth is then combined with the predicted lower resolution depth utilising super-resolution for upsampling similar to \cite{pillai2019superdepth}. The fused depth is then sent to a set of convolutional layers with stride 1, for predicting refined depth. This super-resolved residual architecture induces our model to decipher structurally rich intricate details and refine scene structures while preserving global image layout. We use pixel shuffling after convolving feature with a set of 32, 32, 16, 4 channels to super resolve depth to twice the input resolution.

\begin{table*}
\begin{tabular}{lccccccccc}
\hline
Method                                                                     & Resolution & Dataset & \cellcolor[HTML]{CBCEFB}Abs Rel & \cellcolor[HTML]{CBCEFB}Sq Rel & \cellcolor[HTML]{CBCEFB}RMSE & \cellcolor[HTML]{CBCEFB}RMSE log & \cellcolor[HTML]{9B9B9B}$\delta$\textless$1.25$ & \cellcolor[HTML]{9B9B9B}$\delta$\textless$1.25^2$ & \cellcolor[HTML]{9B9B9B}$\delta$\textless$1.25^3$ \\ \hline
\rowcolor[HTML]{EFEFEF} 
Garg\cite{garg2016unsupervised} cap 50m                                                        & 620 x 188  & K       & 0.169                           & 1.080                          & 5.104                        & 0.273                            & 0.740                                           & 0.904                                             & 0.962                                             \\
Godard\cite{godard2019digging}                                                     & 640 x 192  & K       & 0.115                           & 1.1010                          & 5.164                        & 0.212                            & 0.858                                           & 0.946                                            & 0.974                                            \\
\rowcolor[HTML]{EFEFEF}
GeoNet\cite{yin2018geonet}                                                                     & 416 x 128  & K       & 0.155                           & 1.296                          & 5.857                        & 0.233                            & 0.793                                           & 0.931                                             & 0.973                                             \\
UnDeepVO\cite{li2018undeepvo}                                                                   & 416 x 128  & K       & 0.183                           & 1.730                           & 6.57                         & 0.268                            & -                                               & -                                                 & -                                                 \\
\rowcolor[HTML]{EFEFEF}
Godard\cite{godard2017unsupervised}                                                       & 640 x 192  & K       & 0.148                           & 1.344                          & 5.927                        & 0.247                            & 0.803                                           & 0.922                                             & 0.964                                             \\
GANVO\cite{almalioglu2019ganvo}                                                                       & 416 x 128 & K       & 0.150                                & 0.1141                               & 5.448                             &    0.216                              &    0.808                                             & 0.937                                                  &    0.975                                               \\
\rowcolor[HTML]{EFEFEF}
GLNet\cite{chen2019structure}                                                                       & 416 x 128 & K       & 0.135                                & 1.070                               & 5.230                             &    0.210                              &    0.841                                             & 0.948                                                  &    0.980                                               \\
SuperDepth\cite{pillai2019superdepth}                                                                 & 1024 x 384 & K       & 0.112                           & 0.875                          & 4.958                        & 0.207                            & 0.852                                           & 0.947                                             & 0.977                                             \\
\rowcolor[HTML]{EFEFEF}
UnDispNet\cite{kaushik2019undispnet}                                                                       & 1024 x 384 & K       & 0.110                                & 0.892                               & 4.895                             &    0.206                              &    0.868                                             & 0.951                                                  &    0.976                                               \\
SGANVO\cite{feng2019sganvo}                                                                       & 416 x 128 & K       & 0.065                                & 0.673                               & 4.003                             &    0.136                              &    0.944                                             & 0.979                                                  &    0.991                                               \\
\rowcolor[HTML]{EFEFEF}
Ours                                                                       & 1024 x 384 & K    & 0.960                                & 0.851                               &    4.386                          &   0.179                              &    0.878                                             & 0.962                                                 &    0.984                                              \\ \hline
\end{tabular}
\caption{Self-Supervised depth estimation results on the KITTI dataset \cite{geiger2013vision} using the Eigen Split \cite{eigen2014depth} for depths at cap 80m, as described in \cite{eigen2014depth}.   }
\label{table:tab2}
\end{table*}

\subsection{Loss Function}
Our network predicts depth $D_l$ at 4 scales for left input image $I_l$. Similarly, it predicts depth $D_r$. Provided multi-scale depth of left and right images, we compute appearance matching loss, disparity smoothness loss, left-right consistency loss\cite{godard2017unsupervised}, with occlusion regularization\cite{yang2018deep}. Our losses are computed at 4 scales with standard weight factor at every scale\cite{godard2017unsupervised}.
Our architecture learns a combination of these multi-scale losses in an end-to-end manner. Our fusion network selects features from both upper and lower levels at every level in the encoder network, thereby creating multiple feature pyramid sub-networks that are fed to the decoder after applying the CoordConv solution. Our pyramidal encoder utilises a combination of high level and lower level features, learning global image structure as well as the intricate shape details present in the scene. We also propose a refinement module learning higher scale residual depth from a combination of higher level deep features and lower level residual depth using a pixel shuffling framework that super-resolves lower level residual depth.

\section{EXPERIMENTS}
\label{sec:experiments}
We evaluate our method on KITTI Dataset\cite{geiger2013vision} for both KITTI and Eigen \cite{eigen2014depth} test-train data splits for fair comparison. Our network is trained from scratch with input image at 1080x384 resolution. D1-all and depth metrics from \cite{eigen2014depth},\cite{geiger2013vision} are used for comparison.

\subsection{Implementation Details}
Our base architecture consists of VGG14 model with skip connections. We train our model using Adam optimizer with $10^{-4}$ as learning rate for 70 epochs, where first 50 epochs are trained on all 4 scales, next 10 epochs on 2 scales and last 10 with no regularization and smoothness loss for fine-tuning. Inferencing takes 30ms using NVIDIA RTX 2080Ti GPU. Our network has only 27 million parameters as compared to a basic ResNet 50 having 44 million parameters \cite{godard2017unsupervised} or a stacked module containing 63 million parameters \cite{yang2018deep}.
\subsection{KITTI split}
We tested our method on 200 stereo images provided by KITTI 2015 Stereo Dataset.
Our model preforms drastically well when compared with other methods as shown in Table\ref{table:tab1}. Our architecture outperforms \cite{godard2017unsupervised} which shows the optimality of our architecture in predicting depth. Feature fusion helps in preserving structure and learning refined depth. Super-resolution makes sure that the learnt residual depth isn't inconsistent due to sub-optimal sampling. Also, we observe that feature fusion does significant improvements on depth prediction and CoordConv also gives slight performance boost in depth evaluation.

\subsection{Eigen Split}
As shown in Table \ref{table:tab2}, we observe that our method performs better than rest of the methods exploiting geometric constraints for self-supervised depth prediction. Our method produces visually rich results as show in Fig\ref{fig:fig2}. SGANVO\cite{feng2019sganvo} which utilises stacked generators with adversarial losses performs better but predicts depth at smaller resolution. Our method predicts accurate depth estimates and in future can also utilize GANs for further optimization. We show that architecture has as important role in learning better depth as does having better constraints enforced in form of training loss. Following the same paradigm, we observe that our model performs significantly well than other methods trained on similar losses\cite{godard2017unsupervised,pillai2019superdepth,kaushik2019undispnet,li2018undeepvo}. Compared with the others, our method generates more clearer textures and fine grained details in the predicted depth. 
\section{CONCLUSION}
\label{sec:conclusion}
In this work, we propose a deep feature fusion based architecture leveraging multiple feature pyramids as sub-networks for an optimal encoder network. We combine encoded features with the CoordConv solution thereby learning robust invariant features refined by a residual decoder that incorporates depth super resolution for learning fine-grained depth. Our model predicts accurate depth at higher resolution than other methods. In future, we would like to use adversarial training scheme along with a separate pose network to facilitate learning by monocular video and further improve the performance.   

\bibliographystyle{IEEEbib}
\bibliography{strings,refs}

\end{document}